\documentclass[lettersize,journal]{IEEEtran}

\usepackage{amsmath,amsfonts}
\usepackage{algorithmic}
\usepackage{array}
\usepackage[caption=false,font=normalsize,labelfont=sf,textfont=sf]{subfig}
\usepackage{textcomp}
\usepackage{stfloats}
\usepackage{url}

\usepackage{cite}
\usepackage{multirow}
\usepackage{hyperref}
\usepackage[table]{xcolor}

\usepackage[defaultcolor=red]{changes}

\usepackage{verbatim}
\usepackage{graphicx}
\def\BibTeX{{\rm B\kern-.05em{\sc i\kern-.025em b}\kern-.08em
    T\kern-.1667em\lower.7ex\hbox{E}\kern-.125emX}}
\usepackage{balance}

\begin{document}
\title{Multi-Modal Decouple and Recouple Network for Robust 3D Object Detection}
\author{Rui Ding, Zhaonian Kuang, Yuzhe Ji, Meng Yang, \textit{Member, IEEE}, Xinhu Zheng, \textit{Member, IEEE}, \\ and Gang Hua, \textit{Fellow, IEEE}

\thanks{


\textit{Corresponding authors: M. Yang and X. Zheng}.

R. Ding, Z. Kuang, and M. Yang are with the State Key Laboratory of Human-Machine Hybrid Augmented Intelligence, Institute of Artificial Intelligence and Robotics, Xi'an Jiaotong University, Xi'an, P.R.China. E-mails: \{dingrui, znkwong\}@stu.xjtu.edu.cn; mengyang@mail.xjtu.edu.cn.

Y. Ji and X. Zheng are with the Intelligent Transportation Thrust of the Systems Hub, The Hong Kong University of Science and Technology (Guangzhou), P.R.China. E-mails: yji755@connect.hkust-gz.edu.cn; xinhuzheng@hkust-gz.edu.cn.

G. Hua is with the Multimodal Experiences Research Lab, Dolby Laboratories, Los Angeles, CA, USA. E-mail: ganghua@gmail.com.


}}


\maketitle

\begin{abstract}

Multi-modal 3D object detection with bird’s eye view (BEV) has achieved desired advances on benchmarks. Nonetheless, the accuracy may drop significantly in the real world due to data corruption such as sensor configurations for LiDAR and scene conditions for camera. 
One design bottleneck of previous models resides in the tightly coupling of multi-modal BEV features during fusion, which may degrade the overall system performance if one modality or both is corrupted. 
To mitigate, we propose a Multi-Modal Decouple and Recouple Network for robust 3D object detection under data corruption. 
Different modalities commonly share some high-level invariant features. We observe that these invariant features across modalities do not always fail simultaneously,
because different types of data corruption affect each modality in distinct ways.
These invariant features can be recovered across modalities for robust fusion under data corruption.
To this end, we explicitly decouple Camera/LiDAR BEV features into modality-invariant and modality-specific parts. It allows invariant features to compensate each other while mitigates the negative impact of a corrupted modality on the other.
We then recouple these features into three experts to handle different types of data corruption, respectively, i.e., LiDAR, camera, and both.
For each expert, we use modality-invariant features as robust information, while modality-specific features serve as a complement.
Finally, we adaptively fuse the three experts to exact robust features for 3D object detection. 
For validation, we collect a benchmark with a large quantity of data corruption for LiDAR, camera, and both based on nuScenes. Our model is trained on clean nuScenes and tested on all types of data corruption. Our model consistently achieves the best accuracy on both corrupted and clean data compared to recent models.

\end{abstract}

\begin{IEEEkeywords}
3D object detection, multi-modal, fusion, BEV, robust, data corruption.
\end{IEEEkeywords}

\section{Introduction} \label{section-introduction}

\begin{figure}[t]
    \centering
    \includegraphics[width=0.4\textwidth]{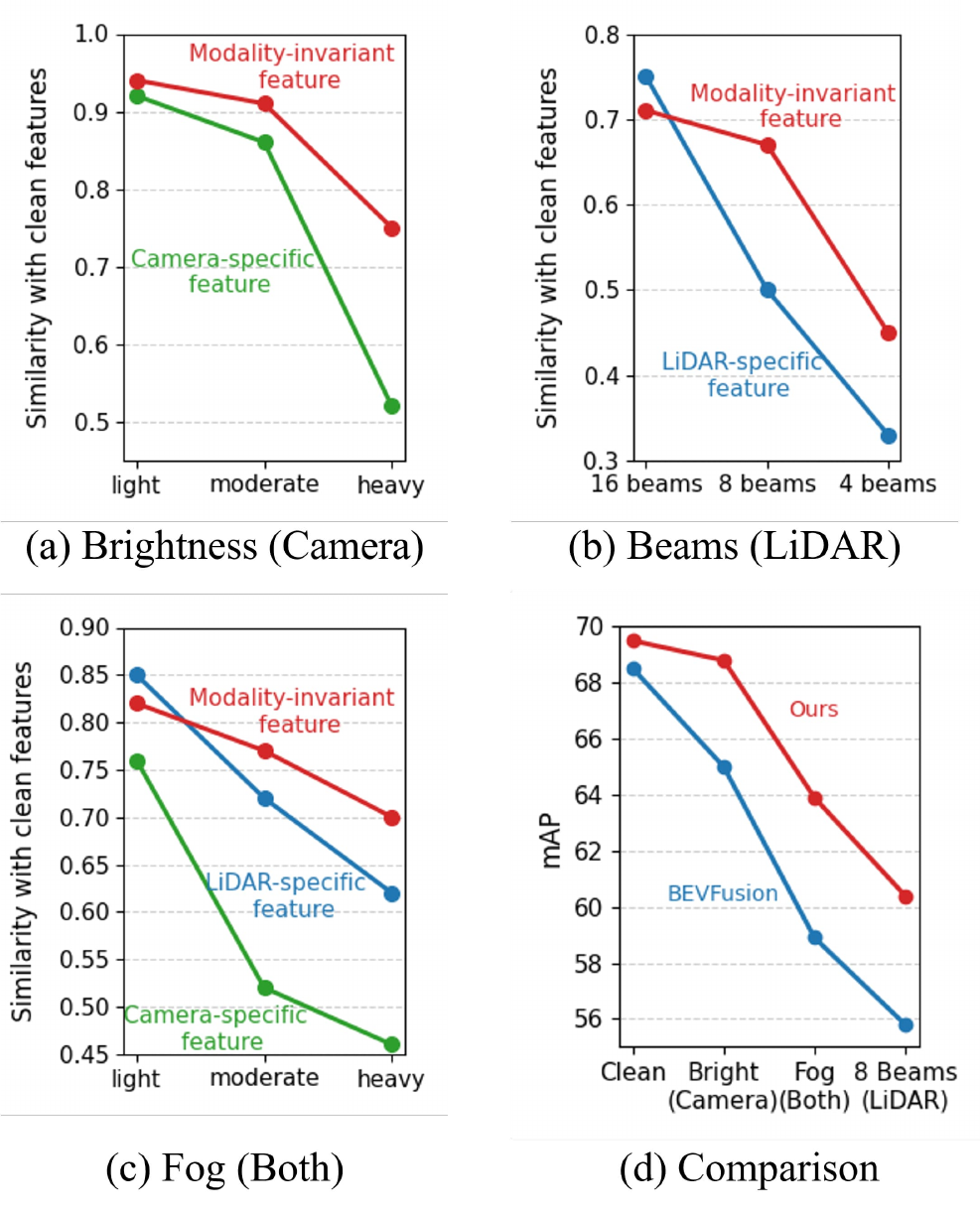}
    \caption{\textbf{Effect of data corruption on decoupled Camera/LiDAR features. }(a)-(c) We measure the similarity (with the metric in \cite{CKA1,CKA2,CKA3}) between clean and corrupted features for modality-invariant and modality-specific parts, respectively. The modality-specific features of LiDAR and Camera are affected seriously under different types of data corruption, while the modality-invariant features are affected slightly. Moreover, modality-invariant features always exhibit higher stability under different corruption severities. These modality-invariant features can be used for robust fusion. (d) Compared to BEVFusion \cite{liu2023bevfusion}, in which LiDAR/Camera BEV features are tightly coupled, our method achieves significant gains on both corrupted and clean data. 
    }
    \label{fig::key_idea}
\end{figure}

\IEEEPARstart{T}{hree}-dimensional (3D) object detection is a crucial technique for planning and navigation of autonomous vehicles. Current models can be categorized into single-modal \cite{wang2021fcos3d,li2021voxel,li2024dualbev,10407035,8897727,9870669,10670218,10693306,10622019} and multi-modal \cite{9826439,10109207,10549844,9693129} based on inputs of LiDAR, camera, or both. Compared to single-modal models, multi-modal ones with both semantics from camera and depth from LiDAR are more promising in the real world. However, the fusion of camera and LiDAR remains a challenge due to their modality gap \cite{yang2022deepinteraction,jiao2023msmdfusion}. 

In recent years, Bird's Eye View (BEV) representation offers an effective way for multi-modal fusion, which uniformly transforms LiDAR and camera inputs into BEV without information loss \cite{liu2023bevfusion}. Therefore, multi-modal 3D detection with BEV \cite{liu2023bevfusion,liang2022bevfusion} has become the dominant solution in both academia and industry. It has achieved desired advances on benchmarks with clean data such as nuScenes \cite{caesar2020nuscenes}.

However, various data corruption may inevitably occur in the real world\cite{9697984}. On the one hand, sensor configurations may differ across vehicles such as reduction of field-of-view (FOV) from 360° to 240°/180°/120°/90°, beams from 32 to 16/8/4/1 in LiDAR, camera views from 6 to 5/3/1, and so on. On the other hand, scene conditions may be affected by severe weather such as rain, fog, snow, and other external disturbances such as crosstalk. More seriously, data corruption may simultaneously occur for both LiDAR and camera. Most multi-modal 3D detection models are still trained and tested on benchmarks with clean data. These types of data corruption may significantly reduce the accuracy of models such as BEVFusion \cite{liu2023bevfusion} in Figure \ref{fig::key_idea}(d).

In recent years, a few models made preliminary attempts to improve the robustness of multi-modal 3D detection under data corruption. A common strategy in current models is to tightly couple features from different modalities during fusion to exploit their complementary information under data corruption. These models generally concatenate features from different modalities \cite{liu2023bevfusion,liang2022bevfusion} or use transformer layers  \cite{ge2023metabev,yang2023crossfusion}. For example, TransFusion \cite{bai2022transfusion} projects one modality to enhance the other, BEVFusion \cite{liu2023bevfusion} employs convolutional fusion, and RobBEV \cite{RobBEV} adopts cross-attention for cross-modal feature interaction.

However, this tightly coupling strategy of multi-modal BEV features during fusion does not always benefit the robustness of current models under data corruption such as the example in Figure \ref{fig::key_idea}(d).
This strategy could fully use the complementary information of both modalities. However, BEV features of two modalities may interfere with each other under data corruption. It even degrades the overall system performance when severe data corruption occurs on either LiDAR, camera, or both.

To mitigate, we propose a novel Multi-Modal Decouple and Recouple network for robust 3D object detection under data corruption. 
To this end, we classify BEV features across modalities into two parts. Modality-invariant features refer to shared information across modalities to describe the same 3D object properties such as category, position, and size. These features can be extracted from image and LiDAR separately. Modality-specific features refer to unique information of each modality such as semantics in images and depth in LiDAR. Removing these features cannot be compensated by the other.

We observe that these invariant features across modalities do not always fail simultaneously even when both modalities are corrupted, because different types of data corruption affect each modality in distinct ways. For example, fog blurs camera images while reduces LiDAR intensity at long distances. It causes each modality to preserve different parts of the invariant features under data corruption. Figure \ref{fig::key_idea}(a)-(c) verifies that these invariant features are affected slightly by data corruption, while the other specific features of LiDAR/Camera are affected. These invariant features can be recovered across modalities for robust fusion under data corruption.

We first develop a modality decouple module to decouple Camera/LiDAR BEV features into modality-invariant and modality-specific features. Modality-invariant features are obtained by first passing LiDAR and camera features into a shared encoder, followed by a similarity loss to enforce invariance. Modality-specific features are extracted by applying an orthogonality loss with invariant features to remove invariant information of modalities. 
This module allows invariant features to compensate for missing information in corrupted modalities, while
mitigating the negative impact of a corrupted modality on the other.

We then develop a modality recouple module to extract robust features for fusion.
We recouple modality-invariant and modality-specific features from LiDAR and camera into three experts to handle different types of data corruption, respectively, i.e., LiDAR, camera, and both. For each expert, we use modality-invariant features as robust information, while modality-specific features serve as a complement. Finally, we adaptively fuse the three experts with a soft weighting mechanism that assigns higher weights to more reliable experts.
 
For validation, we collect a benchmark dataset with a large quantity of data corruption induced by sensor configurations and scene conditions of LiDAR, camera, and both, based on Robo3D \cite{kong2023robo3d} and RoboBEV \cite{xie2023robobev}. Our model is trained on clean nuScenes and tested on different types of data corruption.
Extensive experiments verify that our model consistently achieves the best accuracy on all types of data corruption as well as clean data such as the example in Figure \ref{fig::key_idea}(d).

Our main contributions are highlighted as follows:

\begin{itemize}
\item We observe that invariant features across modalities do not always fail simultaneously when camera, LiDAR, or both are corrupted, which can be used for robust fusion.
\item We propose a robust 3D object detection model by decoupling camera/LiDAR features into invariant and specific parts, recoupling three experts for different types of data corruption, and fusing them to extract robust features.   

\item We collect a benchmark dataset with more types of data corruption.
Our model achieves the best accuracy on all types of data corruption and clean data.
\end{itemize}



\section{Related works} \label{section-realted_works}

\subsection{Multi-modal 3D detection}

Multi-modal 3D detection \cite{vora2020pointpainting,wang2021pointaugmenting,huang2020epnet,li2022deepfusion,liang2022bevfusion} has achieved desired advances compared to single-modal ones \cite{huang2021bevdet,li2023bevdepth,10325411,9792578,10045830,10123008,10214314,10539994}. These models can further be divided into three categories according to the fusion strategies of LiDAR and camera. (a)\textit{ Point-level Fusion}. These models \cite{vora2020pointpainting,wang2021pointaugmenting,huang2020epnet} project image features to raw point clouds of LiDAR, and use image features as complementary information for point clouds. (b) \textit{Feature-level Fusion}. These models \cite{bai2022transfusion,li2022deepfusion,sindagi2019mvx} project image features to LiDAR features space and fuse these features with transformer-based mechanisms. (c) \textit{BEV-level Fusion}. These models \cite{liu2023bevfusion,liang2022bevfusion} transform camera and LiDAR features into a unified BEV view to mitigate the difference of data representations.

BEV-level fusion is currently the mainstream approach. For example, DGFusion \cite{DGFusion} introduces Instance Match Modules to fuse camera and LiDAR features. CMF-IOU \cite{CMF-IoU} employs pseudo point clouds for multi-stage cross-modal fusion. However, current multi-modal models are still trained and tested on benchmarks with clean data. As a result, the overall system performance may drop when tested on corrupted data \cite{dong2023benchmarking,yu2023benchmarking,kong2023robo3d,xie2023robobev,9927485}. In our work, we use BEVFusion \cite{liu2023bevfusion} as the base model and propose a robust multi-modal 3D detection model under data corruption of camera and LiDAR.

\subsection{Data corruption}

In recent years, a few models made preliminary attempts to improve robustness under data corruption. Transfusion \cite{bai2022transfusion} focus on handling severe data corruption of images.  
CrossFusion \cite{yang2023crossfusion} uses a cross-modal complementation strategy and MetaBEV \cite{ge2023metabev} uses a BEV-Evolving decoder block to handle various data corruption of sensors. GraphBEV \cite{song2024graphbev} and Graphalign++ \cite{song2023graphalign++} solves the misalignment between LiDAR and camera BEV features to improve the robustness. RobBEV \cite{RobBEV} proposes a mutual deformable attention and temporal aggregation to handle four types of data corruption of LiDAR and camera. RoboFusion \cite{song2024robofusion} introduces the Segment Anything Model (SAM) \cite{kirillov2023SAM} to enhance robustness.
In addition, data corruption can be mitigated by simulating more types of corrupted data during training \cite{qian2024weatherdg, wu2024imageaug, wu2024prompt} or using semi-supervised methods \cite{chen2025dyconfidmatch}.

Most of current models tightly couple multi-modal BEV features during fusion to effectively utilize complement information across modalities. However, when data corruption occurs on either LiDAR or camera, the two modalities may also interfere with each other. It may degrade the overall system performance on unknown data corruption. In our work, we propose a Multi-Modal Decouple and Recouple network by decoupling Camera/LiDAR BEV features into modality-invariant and modality-specific parts, and adaptively recoupling these features to extract robust features under all types of corruptions.

\begin{figure*}[t!]
    \centering
    \includegraphics[width=1.0\linewidth]{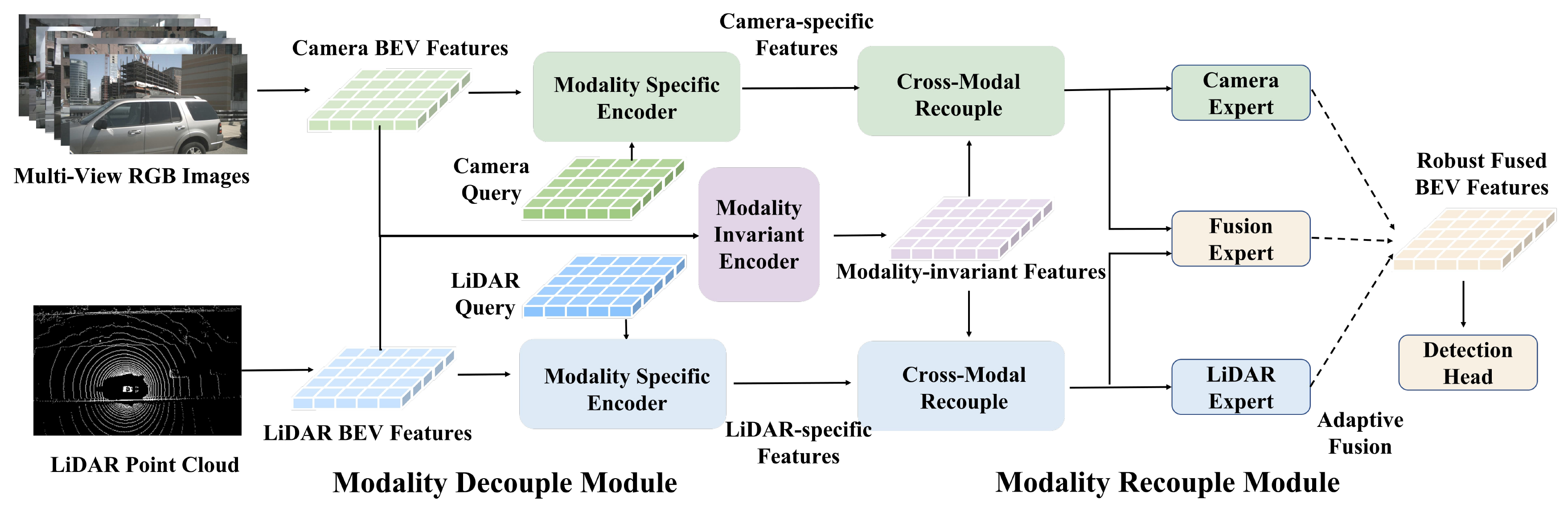}
    \caption{\textbf{Overview of our framework}. First, the LiDAR and camera encoders extract BEV features from point cloud and multi-view images, respectively. Second, \textbf{the Modality Decouple Module }decouples LiDAR/camera BEV features into modality-invariant and modality-specific parts. Then, \textbf{the Modality Recouple Module} recouples the invariant and specific parts into three experts for LiDAR, camera, and fusion, respectively.  Finally, the three experts are adaptively fused to generate robust features for fusion based on corruption level of each modality. The detection heads take the fused features for robust 3D object detection. Our model is trained on clean data, and tested on various types of data corruption in the inference to ensure its robustness in the real world.
} 
    \label{fig:framework}
\end{figure*}

\subsection{Feature-decoupling methods}

In other fields, some work employs cross-modal consistency \cite{hazarika2020misa}, contrastive learning \cite{xia2023achieving}, orthogonality constraints \cite{zhou2023feature}, or mutual-information minimization \cite{xia2023achieving} to obtain modality-invariant features. \cite{zhou2023feature} uses feature decoupling to mitigate negative transfer in multi-task learning.\cite{yu2024pedestrian, yu2024mv} use cross-view invariant feature to improve pedestrian re-identification accuracy. \cite{hazarika2020misa} uses modality-invariant features to achieve effective multi-modal fusion.

The goal of these methods is to improve the fusion accuracy of multi-modal tasks. It is based on the fact that invariant features of different modalities provide common representations for downstream tasks. By comparison, our method aims to improve the robustness of multi-modal fusion under data corruption. It is based on our observation that invariant features of different modalities do not always fail simultaneously, when one modality or both are corrupted.

\subsection{Transformer-based architectures}

Transformer-based architectures \cite{9504551,9438625,10124821,carion2020end} have been dominant for multi-modal 3D object detection in recent years. These solutions often use a set of learnable queries to query and aggregate features based on DETR \cite{carion2020end} or Deformable DETR \cite{zhu2020deformable}. Deformable DETR adds deformable attention to DETR, which significantly reduces training time and spatial complexity. BEVFormer \cite{li2022bevformer}, MetaBEV, and similar works \cite{kim20223d,wang2022detr3d} successfully apply Deformable DETR in 3D object detection by extending the query in BEV view. Similarly, we employ a transformer-based architecture for feature encoding.



\section{Multi-Modal Decouple and Recouple Network}

\subsection{Overall framework}

Figure \ref{fig:framework} shows the overall framework of our Multi-Modal Decouple and Recouple network for robust 3D object detection. 1) The LiDAR and camera encoders extract features from multi-view cameras and LiDAR sensors, respectively, and transform them into BEV representation. The configurations of camera and LiDAR encoders are the same as our base model BEVFusion \cite{liu2023bevfusion}. 2) The Multi-Modal Decouple and Recouple Encoder employs two modules, namely, \textbf{Modality Decouple Module} and \textbf{Modality Recouple Module}, to create robust BEV features, which are introduced in Subsection \ref{subsection_modality_decouple} and Subsection \ref{subsection_modality_recouple}, respectively. 3) The detection heads finally take fused features for 3D object detection, which is introduced in Subsection \ref{subsection_det_head}.

We use the same camera and LiDAR encoder as BEVFusion \cite{liu2023bevfusion}. We represent the camera and LiDAR BEV features as  $ F_{c} \in R^{C\times X \times Y}$ and $ F_{l} \in R^{D\times X \times Y}$, respectively, where $X$ and $Y$ denote the length and width of BEV features, $C$ and $D$ denote the channel dimensions of camera and LiDAR features, respectively.


\subsection{Modality decouple module} 
\label{subsection_modality_decouple}
\begin{figure*}[t]
    \centering
    \includegraphics[width=1.0\linewidth]{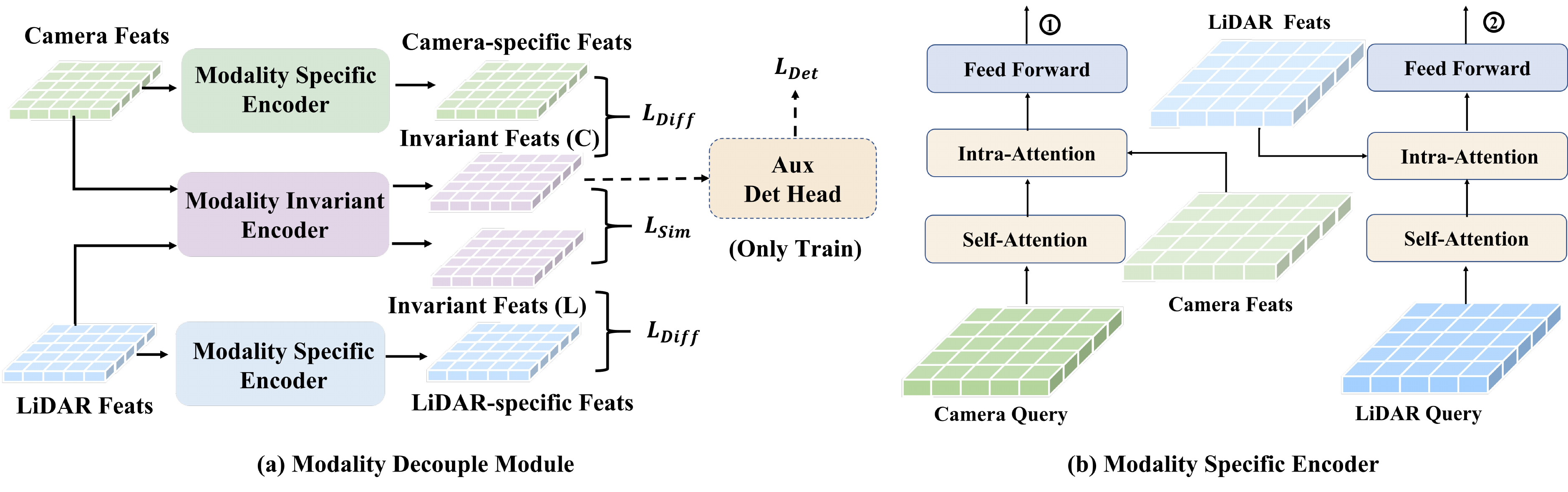}
    \caption{\textbf{Modality Decouple Module}. Three encoders decouple features into modality-invariant and modality-specific parts. The specific features capture information unique to each modality. Modality invariant encoder extracts invariant features across modalities. $L_{Sim}$ enforces consistency output features between camera and LiDAR, while $L_{Diff}$ separates them from modality-specific ones. An auxiliary invariant head, used only during training, ensures the invariant features are truly useful for detection.}
    \label{fig::Modality decouple module}
\end{figure*}

This module is designed to extract robust modality-invariant and modality-specific features, while mitigating the interference of corrupted modality during fusion. As shown in Figure \ref{fig::Modality decouple module}, this module includes two modality-specific encoders for camera and LiDAR, and one modality-invariant encoder. The specific encoders extract features unique to each modality (e.g., camera semantics and LiDAR depth), while the invariant encoder learns invariant features across modalities (e.g., high-level structural or semantic information).

\subsubsection{\textbf{Modality invariant encoder}}
\label{subsubsection_Modality_invariant_encoder}

Invariant features can be extracted from either LiDAR or camera. Even when both modalities are corrupted, different types of data corruption affect them in distinct ways, causing each modality to retain different parts of the invariant features. Therefore, it is robust under data corruption.

To extract modality-invariant features, the LiDAR and camera features are first passed through a shared encoder. Second, a similarity loss is applied between their outputs, together with a difference loss (e.g., the orthogonality loss) between the modality-invariant and modality-specific features. Third, an auxiliary detection head ensures that the learned representation is truly invariant and remains effective for detection. The details are as follows.

In modality-invariant encoder, we first apply one convolution layer to project camera and LiDAR features into the same channel dimension $I$. The projected features are then passed through a shared modality-invariant encoder for both camera and LiDAR. We adopt a simple and efficient two-layer convolution in our implementation. After the encoder, we obtain modality-invariant features $ F_{ic} \in R^{I\times X \times Y}$ and $ F_{il} \in R^{I\times X \times Y}$ from camera and LiDAR, respectively. They refer to invariant feats (C) and invariant feats (L) in Figure \ref{fig::Modality decouple module}.

To ensure the extracted features are truly invariant, we apply a similarity loss $L_{Sim}$ between the invariant features of camera and LiDAR to enforce consistency, while $L_{Diff}$ encourages separate modality-invariant and modality-specific features. The camera and LiDAR features are passed through their respective modality-specific encoders to obtain modality-specific features $F_{sc}$ and $F_{sl}$, respectively. $L_{Sim}$ and $L_{Diff}$ are formulated as:
\begin{equation}
L_{Sim} = \frac{( F_{ic} - F_{il} )^2}{I\times X \times Y} ,  L_{Diff} = \frac{( F_{ic}^T\cdot F_{sc} + F_{il}^T \cdot F_{sl} )}{I\times X \times Y}. 
\end{equation}

Notably, if only above losses are used, the encoder output may collapse to zero values, resulting in low $L_{Sim}$ and $L_{Diff}$ loss values but no meaningful information for detection. To mitigate, we add an auxiliary detection head, which forces the invariant features to detect objects independently. Therefore, it ensures the features are truly useful for 3D object detection. This head shares the same structure as the one in \ref{subsection_det_head} and is only used in training.

\subsubsection{\textbf{Modality specific encoder}}
\label{subsubsection_Modality_specific_encoder}

The modality-specific encoder extracts features unique to each modality, such as camera semantics and LiDAR depth. On the one hand, we design separate encoders for camera and LiDAR to reduce interference from other modalities. On the other hand, deformable attention is used to adaptively extract uncorrupted features, because the input may be partially corrupted. It extracts cleaner features than without decoupling.

To extract modality-specific features, modality queries are first generated by random initialization. Second, these features are fed into separate encoders. Third, a difference loss (e.g., the orthogonality loss) is utilized to remove shared information between the modality-invariant and modality-specific features. The details are as follows.

As shown in Figure \ref{fig::Modality decouple module}(b), our modality specific encoder utilizes a transformer-based architecture, which uses a set of learnable queries to query and aggregate features. Specifically, a single query is decoupled to two independent queries including a camera query $Q_{c}$ and a LiDAR query $Q_{l}$. Both queries consist of two parts: position encoding and embedding. Position encoding is created by assigning each spatial location a coordinate, then processing it through an MLP. The embedding is a randomly initialized $H\times W \times C$ matrix.

Each modality specific encoder contains Self-Attention, Intra-Attention, and Feed Forward layers. The Self-Attention layer learns correlations among queries. The Intra-Attention applies queries for querying and encoding features. The Feed Forward layer updates queries by linear projection. Both the Self-Attention and Intra-Attention layers follow the deformable attention mechanism, which is expressed as follows
\begin{equation}
\small{DeformAttn(q, p, x)=\sum_{i=1}^M \mathcal{W}_i \sum_{j=1}^{\mathrm{N}} \mathcal{A}_{i j} \cdot \mathcal{W}_i^{\prime} x\left(p+\Delta p_{i j}\right)},
\label{deform_attention}
\end{equation}
where $q$, $p$, $x$ represent the queries, reference points, and features, respectively.  $\mathcal{W}_{i}$ and  $\mathcal{W}_i^{\prime}$ are learnable projection matrixes. $M$ and $N$ denote the total number of attention heads and sampled keys, respectively. $i$ and $j$ denote attention head and sampled key indexes. $\mathcal{A}_{i j}$ and  $\Delta p_{i j}$ denote the predicted attention weight and offsets of selected key points, which can be learned as follows
\begin{gather}
\Delta p_{i j}=MLP(q), 
\\
A_{i j}=Softmax(MLP(q)).
\end{gather}

$MLP(\cdot)$ and $softmax(\cdot)$ represent linear projection layer and softmax operator, respectively. Attention weight and offset are both learned from queries, and features are updated by sampling and aggregating. The process of Intra-Attention (IA) for camera and LiDAR can be formulated as follows
\begin{gather}
IA\left(Q_c, F_c\right)=DeformAttn\left(Q_c, p, F_c\right), \label{IA_camra} \\
IA\left(Q_l, F_l\right)=DeformAttn\left(Q_l, p, F_l\right). \label{IA_lidar}
\end{gather}


The query and features are from the same modality in each branch, which ensures that the output queries of the modality decoupling module contain only modality-specific information. This enables the network to mitigate the interference of other modalities. In addition, the learnable sampling offsets and weights help the encoder focus on uncorrupted information, resulting in cleaner features than without decoupling.

\begin{figure*}[t!]
    \centering
    \includegraphics[width=1.0\linewidth]{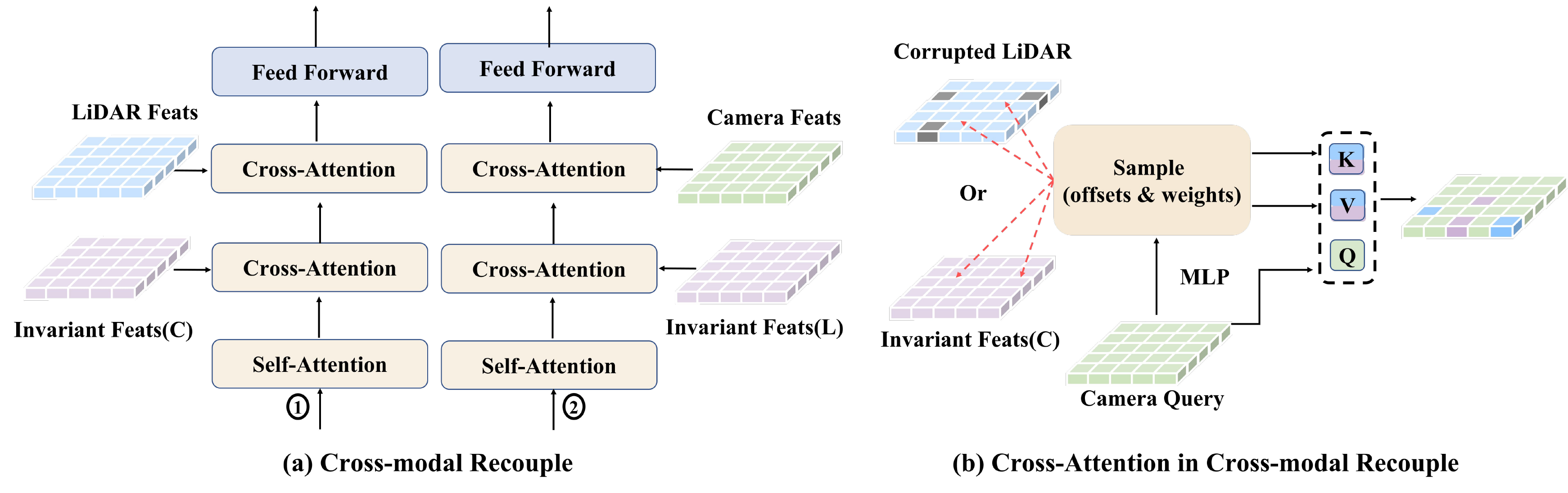}
    \caption{\textbf{Cross-modal recouple in the modality recouple module}. The cross-modal recouple consists of one self-attention and two cross-attention. In cross-attention, the sample offsets and weights are learnable, therefore cross-attention can dynamically extract useful information from corrupted and invariant features as supplements.}
    \label{fig:recouple_module}
\end{figure*}

\subsection{Modality recouple module}
\label{subsection_modality_recouple}

The modality decouple module in last section well decouple modality-invariant and modality-specific features. However, it does not exploit the complementary information between modalities. Therefore, we recouple these features into three experts to handle different types of data corruption, and adaptively fuse them to extract robust features.

\subsubsection{\textbf{Cross-modal recouple}}
\label{subsection_query_enhance}

 Multi-modal LiDAR and camera provide complementary information for 3D object detection even though data corruption occurs in a single or both modalities. And invariant features remain robust under any corruption and can compensate for missing information in corrupted modalities. Therefore, we design a cross-modal recouple to further enhance clean modality with useful information from corrupted and invariant ones. 

Specifically, as shown in Figure \ref{fig:recouple_module}, queries can dynamically sample and aggregate features from the corrupted modality and invariant features by employing a deformable attention mechanism in Equation (\ref{deform_attention}). During this process, the key points and weights for sampling are learned from the queries. It makes full use of valid information in corrupted modality and robust invariant features. Therefore, we could enhance the features of clean modality with the help of complementary information in corrupted and invariant ones. Even though both modalities are corrupted in real world, two corrupted modalities enhance their features with each other. And invariant features can further compensate it.

The cross-modal recouple consists of one Self-Attention, two Cross-Attention, and Feed Forward layers. The first cross-modal attention is designed to add invariant information from modality-invariant features, which remains robust under any modality corruption. The process of Cross-Attention (CA)  can be formulated as follows
\begin{gather}
CA\left(Q_c, F_{il}\right)=DeformAttn\left(Q_c, p, F_{il}\right),
\\
CA\left(Q_l, F_{ic}\right)=DeformAttn\left(Q_l, p, F_{ic}\right).
\end{gather}

The second cross-modal attention project features of corrupted modality to clean one. The process of Cross-Attention (CA) for camera and LiDAR can be formulated as follows
\begin{gather}
CA\left(Q_c, F_l\right)=DeformAttn\left(Q_c, p, F_l\right),
\\
CA\left(Q_l, F_c\right)=DeformAttn\left(Q_l, p, F_c\right).
\end{gather}

With these operations, the cross-modal recouple makes full use of valid information in corrupted modality and invariant features to enhance clean one. The output is enhanced camera feature $F_{c}$ and enhanced LiDAR feature $F_{l}$, each specialized in handling different corruption scenarios.

\subsubsection{\textbf{Adaptive fusion}}
\label{subsection_moe_recouple}

Because the reliability of each enhanced features varies across scenarios, we design the adaptive fusion to dynamically fuse features based on the corruption level of each modality, enabling robust fusion.

Specifically, we design three detection experts $E_c$, $E_l$ and $E_f$ in Figure \ref{fig:framework} to adaptively handle different corruption scenarios. These experts process the enhanced camera feature $F_{c}$, the enhanced LiDAR feature $F_{l}$, and their concatenation $[F_c,F_l]$ respectively. Because the camera and LiDAR features are more reliable when their own modality is less corrupted, and their fusion becomes more effective when both modalities are corrupted, these experts enable the model to adaptively handle different corruption scenarios. The output features are denoted as follows
\begin{equation}
F_{ec} = E_c(F_c), \quad F_{el} = E_l(F_l), \quad F_{ef} = E_f([F_c, F_l]).
\end{equation}

Each expert uses different modality inputs to make them more distinct. Therefore, their performance varies under different corruptions. We adopt a soft fusion strategy to achieve stronger robustness. A lightweight router with two convolutions takes the concatenated features as input and predicts a set of soft weights via a softmax operation:
\begin{equation}
W = Softmax(Router([F_c, F_l])) = [W_{ec}, W_{el}, W_{ef}].
\end{equation}

The final BEV feature is computed as the weighted sum of all expert outputs:
\begin{equation}
F_{out} = W_{ec} \cdot F_{ec} + W_{el} \cdot F_{el} + W_{ef} \cdot F_{ef}.
\end{equation}

In addition, we add an entropy regularization loss to make the outputs of each expert more distinct:
\begin{equation}
L_{reg} = -{\textstyle\sum_{i \in \{ec, el, ef\}} W_i \log W_i}.
\end{equation}

Overall, the adaptive fusion enables the model to adaptively fuse features and assigns higher weights to more reliable experts. It significantly improves the robustness of 3D object detection under data corruption.

\subsection{Detection head}
\label{subsection_det_head}

We employ a Transfusion head to predict the size, center, and rotation of 3D objects. The configuration of the detection head and the losses are consistent with \cite{liu2023bevfusion}. More details are provided in Subsection \ref{subsection_implement_details}.

\begin{figure*}[t]
    \centering
    \includegraphics[width=18cm]{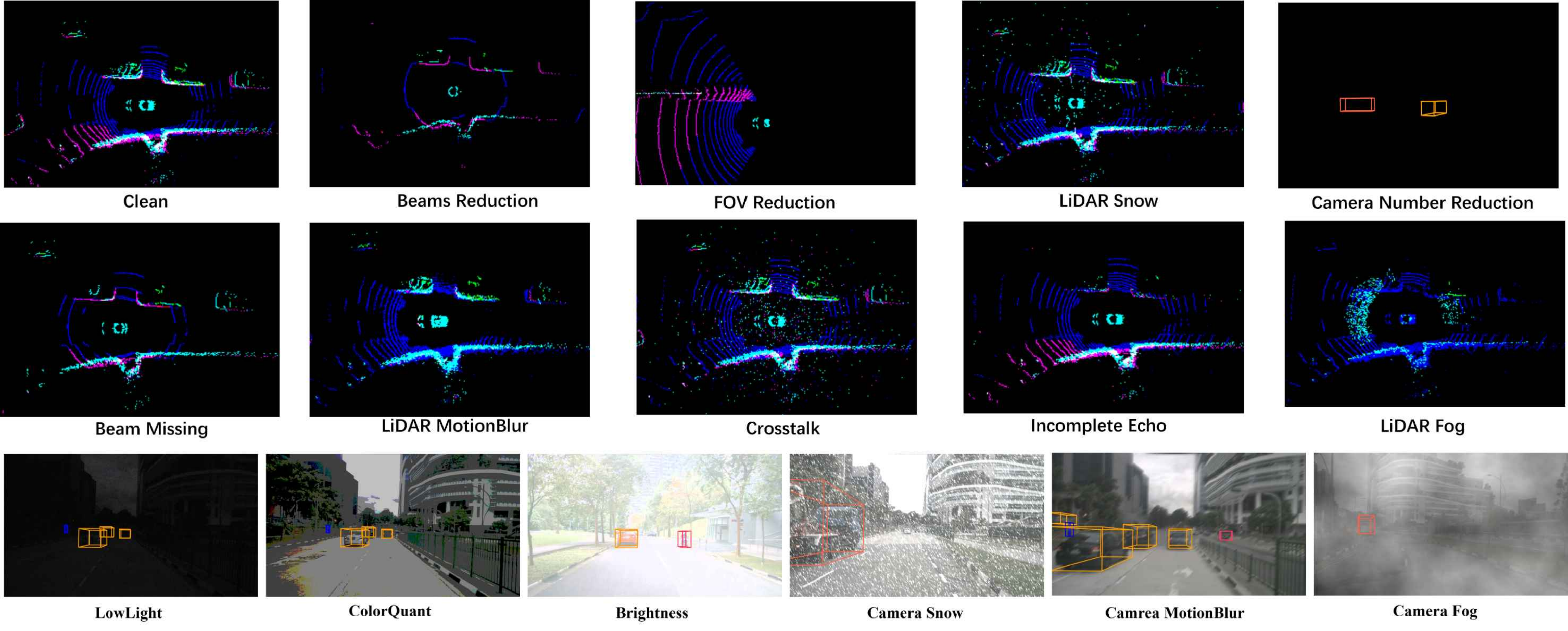}
    \caption{\textbf{Examples of data corruption}. We simulate various types of data corruption on cameras, LiDAR and both in real world, and add them to clean nuScenes validation images to generate corrupted data for testing.}
    \label{fig:detection_vis_demo}
\end{figure*}

\section{Experiments}

\subsection{Implementation details}
\label{subsection_implement_details}

\paragraph{\textbf{Dataset}}

We use the autonomous driving dataset nuScenes \cite{caesar2020nuscenes} as the base data, which is captured with six cameras and a 32-beam LiDAR in 360-degree Field of View (FOV). The dataset is divided into 700 training, 150 validation, and 150 testing scenes. This dataset has clean images from multi-view cameras and point clouds from LiDAR. In recent years, researchers \cite{kong2023robo3d,xie2023robobev,ge2023metabev} have simulated different types of data corruption based on nuScenes. 

In our work, we collect a large-scale test dataset with much more types of real-world data corruption based on these datasets. \textbf{(I) Sensor corruptions} include Beam Reduction from 32 to 16/8/4/1, Field of View (FOV) reduction from 360° to 240°/180°/120°/90°/60°, and Camera Number reduction from 6 to 1/3/5. \textbf{(II) Scene corruptions of LiDAR} include \textit{LiDAR Fog}, \textit{LiDAR Snow}, \textit{LiDAR MotionBlur}, \textit{Beam Missing}, \textit{Crosstalk}, and \textit{Incomplete Echo} based on Robo3D \cite{kong2023robo3d}. \textbf{(III) Scene corruptions of camera} include \textit{Brightness}, \textit{ColorQuant}, \textit{Camera Fog}, \textit{Camera MotionBlur},\textit{Camera Snow}, and \textit{LowLight} based on RoboBEV \cite{xie2023robobev}. Each type of data corruption in \textbf{(II)} and \textbf{(III)} is further categorized into three severity levels: light, moderate, and heavy. \textbf{(IV)  Multi-modal Corruptions} include snow, fog, and motion blur in both LiDAR and camera. Camera and LiDAR are corrupted with the same severity level in each scenario. Notably, current methods only simulate data corruption of either LiDAR or cameras. However, camera and LiDAR may be simultaneously corrupted by severe weather conditions in real-world applications. We finally simulate various types of data corruption in Figure \ref{fig:detection_vis_demo}, and add them to clean nuScenes validation to generate corrupted data for testing.


\paragraph{\textbf{Metrics}}

We use nuScenes Detection Score (NDS) and mean Average Precision (mAP) as the metrics. In addition, we employ the mean Resilience Rate (\textbf{mRR}) \cite{kong2023robo3d} to evaluate the average performance of $N$ types of data corruption as $mRR = \frac{1}{N} \sum_{N}^{i=1} {mAP_{i}}/{mAP_{clean}}$, where  $mAP_{clean}$ and $mAP_{i}$  refer to mAP values on clean data and corrupted data, respectively.


\begin{table*}[t]
    \centering
    \caption{\textbf{Results on sensor corruptions}. 'Clean' refers to the results in nuScenes benchmark. 'CN' refers to camera numbers. The best results are in bold. Our model consistently achieves the best results, especially on severe data corruptions such as 1 beam.}
    \setlength{\tabcolsep}{3.0mm}
    \begin{tabular}{c|ccccccccccc|c}

            \hline 
         
         \multirow{2}{*}{mAP}   & \multicolumn{2}{c|}{Clean} &  \multicolumn{4}{c|}{Beam} &  \multicolumn{4}{c|}{FOV} &  \multicolumn{1}{c|}{CN} & \multirow{2}{*}{mRR}  \\ 
        
        \cline{2-12}
         
        &  Test & \multicolumn{1}{c|}{Val} & 1 & 4 & 8 & \multicolumn{1}{c|}{16} & $90^{\circ}$ & $120^{\circ}$ & $180^{\circ}$ & \multicolumn{1}{c|}{$240^{\circ}$}  & \multicolumn{1}{c|}{5}   \\
        
        \hline

FUTR3D\cite{chen2023futr3d} &69.4 &	67.4 &	11.9 &	44.0	 &49.0	 &50.0	 &19.2 &	24.1	 &32.2	 &40.7	 &46.8 &	52.4$\%$\\

UVTR\cite{li2022UVTR} & 67.1&	65.4&	12.7	&43.7&	52.2	&53.3&	18.6&	23.3&	30.8&	38.8&	57.4&	56.2$\%$\\

Transfusion\cite{bai2022transfusion}  & 68.9 & 67.3 & 4.7 & 39.2 & 49.9 & 51.5 & 14.7 & 21.2 & 29.2 & 39.6 &   61.4& 51.4$\%$ \\

AD$\_$BEVFusion\cite{liang2022bevfusion}   & 69.2 & 67.9 & 8.7 & 43.0 & 52.8 & 54.5 & 15.4 & 21.0 & 31.1 & 39.6 & 59.1 &  53.2$\%$ \\

BEVFusion\cite{liu2023bevfusion}   & 70.2 & 68.5 & 11.3 & 47.3 & 55.8 & 56.9 & 16.4 & 22.0 & 31.0 & 40.3 & 64.1 &  55.9$\%$ \\
 
MetaBEV\cite{ge2023metabev}   & 70.0 & 68.6 & 11.3 & 47.1 & 55.9 & 57.6 & 17.8 & 23.3 & 31.9 & 40.9 & 64.4 & 56.7$\%$ \\

\textbf{Ours} & \textbf{70.5}&  \textbf{69.5} & \textbf{22.8} & \textbf{55.3} & \textbf{60.4} & \textbf{61.5} & \textbf{24.6} & \textbf{29.6} & \textbf{36.9} & \textbf{44.3} & \textbf{65.3} & \textbf{64.1}$\textbf{\%}$\\

\rowcolor{gray!20}
Improvement & +0.5 & +0.9 & +10.1 & +8.0 & +4.5 & +3.9 & +5.4 &  +5.5 & +4.7 & +3.4 & +0.9 & +7.4$\%$ \\

\hline
\end{tabular}
    \label{table::sensor_corruptions}
\end{table*}




\begin{table*}[t]
    \centering
    \caption{\textbf{ Results on scene corruptions of LiDAR}.  We show the results of heavy severity levels.  Our model consistently achieves the best results on all these types of scene corruptions.}
    \setlength{\tabcolsep}{4.0mm}

\begin{tabular}{c|cccccc|c}

\hline 

\multirow{2}{*}{NDS/mAP} & Beam  & \multirow{2}{*}{Crosstalk}  & \multirow{2}{*}{Fog}  & Incomplete  & Motion  & \multirow{2}{*}{Snow}   & \multirow{2}{*}{mRR} \\

& Missing & & & Echo & Blur &  & \\

\hline

FUTR3D\cite{chen2023futr3d} &69.1/64.1	&63.5/56.4	&66.7/60.6	&69.1/64.2	&59.2/57.3	&67.1/61.4	&90.0$\%$ \\

UVTR\cite{li2022UVTR} &68.6/61.3	&64.8/56.9	&64.8/56.2	&67.5/61.3	&60.3/54.7	&66.3/59.2	&89.0$\%$\\

TranFusion\cite{bai2022transfusion}  & 69.5/64.8 & 63.2/54.8 & 66.1/59.4 & 69.4/65.0 & 57.1/53.1 & 67.8/61.5  & 88.9$\%$ \\

AD$\_$BEVFusion\cite{liang2022bevfusion}  & 69.6/65.6 & 62.6/53.8 & 65.9/59.9 & 69.6/65.9 & 58.6/55.5 & 67.5/62.0  & 89.0$\%$ \\

BEVFusion\cite{liu2023bevfusion} & 70.2/66.8 & 65.2/58.4 & 67.9/63.3 & 70.3/67.0 & 59.2/57.6 & 68.9/64.8  & 91.9$\%$ \\

MetaBEV\cite{ge2023metabev}  & 70.2/66.9 & 65.3/58.9 & 67.4/62.8 & 70.3/67.1 & 59.3/57.9 & 68.9/64.7  & 91.9$\%$ \\

\textbf{Ours}  &  \textbf{71.3/68.3} & \textbf{67.6/62.5} & \textbf{69.5/65.6} & \textbf{71.3/68.4} & \textbf{62.7/61.9} & \textbf{70.2/66.5} &  \textbf{94.3}$\textbf{\%}$\\

\rowcolor{gray!20}
Improvement & +1.1/+1.4 & +2.3/+3.6 & +1.6/+2.3 & +1.0/+1.3 & +3.4/+4.0 & +1.3/+1.7 & +2.4$\%$ \\

\hline
\end{tabular}
    \label{table::lidar_scene_corruptions}
\end{table*}

\begin{table*}[t]
    \centering
    \caption{\textbf{ Results on scene corruptions of camera}. We show the results of heavy severity levels. Our model consistently achieves the best results on all these types of scene corruptions. * refers to results from RoboBEV\cite{xie2023robobev}.}
    \setlength{\tabcolsep}{4.0mm}

\begin{tabular}{c|cccccc|c}

\hline 

\multirow{2}{*}{NDS/mAP} & \multirow{2}{*}{Brightness}  & \multirow{2}{*}{ColorQuant}  & \multirow{2}{*}{Fog}  & \multirow{2}{*}{LowLight}  & Motion  & \multirow{2}{*}{Snow}   & \multirow{2}{*}{mRR} \\

&  & & &  & Blur &  & \\

\hline

AutoAlignV2*\cite{AutoAlignV2} &60.8/55.0	&60.1/54.1	&-&	57.8/49.5&	59.0/51.9&	- & -	\\

FUTR3D\cite{chen2023futr3d} &69.6/65.2	&66.2/58.6	&69.3/64.6	&63.7/54.4	&59.1/46.2	&57.1/43.4&	82.1$\%$\\

UVTR\cite{li2022UVTR}& 69.3/63.7&	66.4/58.8	&68.8/62.8	&65.6/57.1	&65.9/57.7	&65.9/57.6&	91.1$\%$\\

Tranfusion\cite{bai2022transfusion}  & 68.2/64.5  & 68.5/63.4 & 68.3/64.5 & 66.4/61.4 & 67.8/63.2 & 67.6/63.0 & 94.1$\%$ \\

AD$\_$BEVFusion\cite{liang2022bevfusion}  & 69.3/65.6 & 68.5/64.1 & 69.3/65.4 & 67.1/61.6 & 67.7/63.3 & 68.2/63.4 & 94.1$\%$ \\

BEVFusion\cite{liu2023bevfusion}  & 69.3/65.0 & 69.0/64.6 & 69.9/65.9 & 67.5/62.0 & 68.8/64.2 & 68.1/63.0 & 93.6$\%$ \\

MetaBEV\cite{ge2023metabev}  & 69.3/65.4 & 69.1/64.8 & 69.9/66.0 & 67.1/61.2 & \textbf{69.3/64.9} & 68.3/63.3 & 93.7$\%$ \\

\textbf{Ours}  & \textbf{71.4/68.8} & \textbf{69.4/65.5} & \textbf{71.2/68.4} & \textbf{67.8/62.3} & 69.1/64.9 & \textbf{68.6/64.1} & \textbf{94.5}$\textbf{\%}$\\

\hline
\end{tabular}
    \label{table::camera_scene_corruptions}
\end{table*}

\begin{table*}[t]
    \centering
    \caption{\textbf{Results on Multi-modal Corruptions}. Our model consistently performs the best when cameras and LiDAR are simultaneously corrupted by heavy fog, snow, and motion blur.}
    \setlength{\tabcolsep}{7.0mm}

\begin{tabular}{c|cccc|c}
\hline 
NDS/mAP  & Clean(val) & Fog & Snow & MotionBlur & mRR \\
\hline 

FUTR3D\cite{chen2023futr3d} & 70.9/67.4	&66.0/59.6	&50.5/32.5	&47.0/34.7	&62.7$\%$ \\

UVTR\cite{li2022UVTR} & 70.2/65.4	&65.2/56.1	&61.7/49.9	&53.9/43.9	&76.4$\%$  \\

TransFusion\cite{bai2022transfusion} & 71.2/67.3 & 64.8/58.0 & 64.0/55.9 & 54.5/50.9 & 81.6$\%$ \\

AD$\_$BEVFusion\cite{liang2022bevfusion} & 71.0/67.9 & 65.1/58.1 & 64.9/56.7 & 55.5/50.6 & 81.1$\%$ \\

BEVFusion\cite{liu2023bevfusion} & 71.4/68.5 & 65.2/58.9 & 64.3/56.4 & 55.6/50.8 & 81.5$\%$ \\

MetaBEV\cite{ge2023metabev} & 71.3/68.6 & 65.6/59.6 & 64.5/56.5 & 55.9/51.4 & 81.4$\%$ \\

\textbf{Ours} & \textbf{72.0/69.5} & \textbf{68.6/63.9} & \textbf{65.0/58.4 }& \textbf{57.4/52.8} & \textbf{84.0}$\textbf{\%}$\\

\hline
\end{tabular}

    \label{table::multi_modal_corruptions}
\end{table*}



\paragraph{\textbf{Network configurations}}

We use BEVFusion \cite{liu2023bevfusion} as our base model. Camera images are downsampled to 256×704 and LiDAR point clouds are voxelized at 0.075m. Swin Transformer \cite{liu2021swin} and VoxelNet \cite{zhou2018voxelnet} are utilized as the backbone for image and LiDAR, respectively. The channel dimensions are set to 80 for camera query and 256 for LiDAR query. The BEV grid shape is set as 180×180 for both queries and BEV feature maps.


\paragraph{\textbf{Training details}}

Our model is trained on clean nuScenes and tested on unknown corruptions. We employ the AdamW optimizer with a learning rate of 5e-5. Our model is trained overall 6 epochs after loading pre-trained camera and LiDAR encoder model \cite{mmdet3d2020}. Notably our model can get the on-par performance when trained 24 epochs without loading pre-trained models. All models are trained on 4 RTX3090 GPUs, and evaluated on 1 RTX3090 GPU. Notably, we do not apply Test-Time Augmentation (TTA) and model ensembles.

\subsection{Evaluation}

First of all, We train our model and the baselines on clean nuScenes and comprehensively test their robustness on all types of unknown data corruption in the collected test dataset including sensor corruptions \textbf{(I)}, scene corruptions \textbf{(II)(III)}, and multi-modal corruptions \textbf{(IV)}. Because some recent models are trained and tested under known data corruption. We additionally conduct experiments following the same setting, for fair comparison with them.

\subsubsection{\textbf{Sensor corruptions}}

Table \ref{table::sensor_corruptions} shows the mAP results of all models on sensor corruptions, including reduction of beam, FOV, and cameras. First, our model achieves the best result on both validation and test sets of clean nuScenes benchmark. It demonstrates that the proposed method well improves the robustness on data corruption and also improve the performance on clean data. Second, our model consistently achieves the best results on all types of sensor corruptions. Moreover, our model often achieves more significant gains on severe data corruption such as FOV from 360° to 90° and beams from 32 to 1 in LiDAR. In total, the results verify the robustness of our model on all types of unknown sensor corruption.

\subsubsection{\textbf{Scene corruptions}}

We test all models on scene corruptions of both camera and LiDAR. For clarity, Table \ref{table::lidar_scene_corruptions} and \ref{table::camera_scene_corruptions} show the results of scene corruptions (at high severity levels) on LiDAR and camera, separately. Our model also consistently achieves the best results on all these types of unknown scene corruptions. Notably, the accuracy of all models drops significantly on data corruption of LiDAR than those of camera. It indicates that LiDAR plays a crucial role in multi-modal 3D detection. In total, the results verify the robustness of our model on all types of unknown scene corruptions of both LiDAR and camera in real-world applications.



\begin{table}[t]
    \centering
    \caption{\textbf{Comparison under known corruptions}. * refers to results in RobBEV\cite{RobBEV}.}
    \setlength{\tabcolsep}{1.5mm}

\begin{tabular}{c|ccc}
\hline 
Method  & Clean(val) & FOV $180^{\circ}$ & FOV $120^{\circ}$ \\
\hline
Transfusion*\cite{bai2022transfusion}        & 70.9/66.9 & 51.4/29.3 & 45.8/20.3\\
BEVFusion*\cite{liu2023bevfusion}            & 71.0/67.9 & 55.8/46.4 & 50.8/41.5\\
RobBEV*\cite{RobBEV}                         & 71.4/68.7 & 57.9/48.2 & 54.0/43.3\\  
MetaBEV\cite{ge2023metabev}  & 71.3/68.6 & 56.7/47.8 & 52.6/42.7 \\
\textbf{Ours }  & \textbf{72.0/69.5} & \textbf{59.5/49.6} & \textbf{55.3/45.2} \\

\hline
\end{tabular}

    \label{table::compare_in_domain}
\end{table}

\begin{table}[t]
    \centering
    \caption{\textbf{Comparison on the nuScenes validation set}.}
    \setlength{\tabcolsep}{5.0mm}

\begin{tabular}{c|c}
\hline 
Method  & NDS/mAP  \\
\hline
BEVFusion\cite{liu2023bevfusion}           & 71.4/68.5 \\
MetaBEV\cite{ge2023metabev}  & 71.3/68.6  \\
RoboFusion\cite{song2024robofusion} & 71.3/69.1\\
RobBEV\cite{RobBEV}                         & 71.4/68.7 \\
SyNet\cite{SyNet} & 72.2/69.4 \\
DGFusion\cite{DGFusion}  & 71.9/69.2 \\
CMF-IoU\cite{CMF-IoU}    & 72.4/69.1\\
\textbf{Ours }  & \textbf{72.5/69.8}  \\

\hline
\end{tabular}

    \label{table::compare_recent methods}
\end{table}


\subsubsection{\textbf{Multi-modal corruptions}}

Table \ref{table::multi_modal_corruptions} shows the results of all models on multi-modal corruptions, i.e. camera and LiDAR are simultaneously corrupted by heavy fog, snow, and motion blur. It is seen that the accuracy of all these models drops more significantly on multi-modal corruptions than on data corruption of a single modality in Table \ref{table::lidar_scene_corruptions} and Table \ref{table::camera_scene_corruptions}. This is because these models can still use information from clean modality when only one modality is corrupted. However, these models may fail to utilize complementary information of both two modalities when they are corrupted simultaneously. Table \ref{table::multi_modal_corruptions} shows that our model still consistently achieves the best results on all these types of multi-modal corruptions compared to the baseline models. It is mainly attributed to the design of our decouple and recouple module. When both modalities are corrupted, our model adaptively fuses information to maintain the accuracy of multi-modal 3D detection.

\subsubsection{\textbf{Comparison in known corruptions}}

Both our model and recent models are trained under simulated FOV-limited settings for a fair comparison. Table \ref{table::compare_in_domain} shows our model is more robust on known corruptions. It is because our decouple module mitigates the negative impact of a corrupted modality on the other, while the invariant features further provide more robust information. The recouple module focuses more on the camera expert during fusion. Overall, our model achieves stronger robustness under both known and unknown corruptions.

\subsubsection{\textbf{Comparison on nuScenes validation set}}

We additionally compare our method with recent representative multi-modal 3D detection models \cite{song2024robofusion, SyNet, DGFusion, CMF-IoU} on nuScenes validation set with clean data. Table \ref{table::compare_recent methods} shows that our model still achieves the best result compared to these methods. It further verifies the effectiveness of our method on not only corrupted data but also clean data.


\begin{table}[t]
\centering
\caption{\textbf{Effectiveness of decouple and recouple modules.} "decouple" and "recouple" refer to the modality decouple module and the modality recouple module, respectively, in our network.
}
\setlength{\tabcolsep}{2.5mm}
\begin{center}
\begin{tabular}{c|cc|cc}
\hline
\# & decouple  & recouple & NDS/mAP & mRR \\
\hline
a  &                    &          &    71.0/68.5  & 77.4$\%$   \\
b  &\checkmark               &          &  71.3/68.6 & 78.2$\%$ \\
d  &                         &\checkmark&    71.8/69.1 & 78.9$\%$ \\
\rowcolor{gray!20}
e  &\checkmark& \checkmark &   \textbf{72.0/69.5}      &  \textbf{81.7}$\textbf{\%}$ \\
    \hline
    \end{tabular}
    \label{tab:ablations_our_modules}
\end{center}
\end{table}

\begin{table}[t]
\centering
\caption{\textbf{Ablation study of modality specific encoder.} “None” refers to not using the modality-specific encoder.
}
\setlength{\tabcolsep}{2.7mm}
\begin{center}
\begin{tabular}{c|c|cc}
\hline
\# & modality specific encoder  & NDS/mAP & mRR \\
\hline
a  &    none            & 71.0/68.5  & 77.4$\%$ \\
b  & two CNN layers  & 71.4/68.6  & 77.4$\%$ \\
\rowcolor{gray!20}
c  &  deformable attention     &  \textbf{71.7/69.2} & \textbf{77.9}$\%$ \\

    \hline
    \end{tabular}
    \label{tab:ablations_specific_encoder}
\end{center}
\end{table}

\begin{table}[t]
\centering
\caption{\textbf{Ablation studies of modality invariant encoder.} 
}
\setlength{\tabcolsep}{2.0mm}
\begin{center}
\begin{tabular}{c|ccc|cc}
\hline
\# & $L_{Sim}$ & $L_{Diff}$ & Aux head  & NDS/mAP & mRR \\
\hline
a  &          &           &         &                   71.0/68.5  & 77.4$\%$ \\
b  & \checkmark  &  \checkmark     &     &      70.3/67.6   &    74.2$\%$               \\
c  &           &           & \checkmark&           70.7/68.2        & 77.0$\%$\\
\rowcolor{gray!20}
d  &\checkmark&\checkmark & \checkmark&       \textbf{71.3/68.6} & \textbf{78.2}$\textbf{\%}$  \\
    \hline
    \end{tabular}
    \label{tab:ablations_invariant}
\end{center}
\end{table}
\begin{table}[t]
\centering
\caption{\textbf{Ablation studies of cross-modal recouple.} 
}
\setlength{\tabcolsep}{2.0mm}
\begin{center}
\begin{tabular}{c|cc|cc}
\hline
\# & other modality & invariant  & NDS/mAP & mRR \\
\hline
a  &                &            &    71.0/68.5  & 77.4$\%$ \\
b  & \checkmark     &            &    71.2/68.5 & 77.9$\%$\\
c  &                & \checkmark &    71.5/68.9 & 78.5$\%$      \\
\rowcolor{gray!20}
d  & \checkmark     & \checkmark &      \textbf{71.8/69.3}      & \textbf{79.4}$\textbf{\%}$      \\
    \hline
    \end{tabular}
    \label{tab:ablations_enhance}
\end{center}
\end{table}
\begin{table}[t]
\centering
\caption{\textbf{Ablation studies of adaptive fusion.} “diff input” indicates that each expert receives different modality features as input.
}
\setlength{\tabcolsep}{1.5mm}
\begin{center}
\begin{tabular}{c|ccc|cc}
\hline
\# & diff input & soft recouple & entropy loss  & NDS/mAP & mRR \\
\hline
a  &          &           &         & 71.0/68.5  & 77.4$\%$\\
b  & \checkmark &         &         & 71.7/68.6  & 77.6$\%$\\
c  &           & \checkmark &       & 71.7/68.7  & 78.7$\%$\\
d  &           &           & \checkmark & 71.1/68.6  & 77.5$\%$\\
\rowcolor{gray!20}
e  & \checkmark &\checkmark & \checkmark&   \textbf{71.8/69.1} & \textbf{78.9}$\textbf{\%}$  \\
    \hline
    \end{tabular}
    \label{tab:ablations_moe_recouple}
\end{center}
\end{table}

\begin{table}[t]
\centering
\caption{\textbf{Layers of decouple and recouple modules.} 
}
\setlength{\tabcolsep}{5.0mm}
\begin{center}
    \begin{tabular}{c c c}
    \hline
    Layers    & NDS/mAP & mRR  \\
    \hline
        0 & 71.0/68.5  & 77.4$\%$\\
        \rowcolor{gray!20}
        \textbf{1MD+1MR} & \textbf{71.5}/\textbf{68.9} & 78.5$\%$ \\
        1MD+2MR & 70.9/68.0  & \textbf{78.9}$\%$\\
        2MD+2MR & 71.2/68.5 & 77.9$\%$\\
    \hline
    \end{tabular}
    \label{tab:ablations_modules_layer}
\end{center}
\end{table}

\begin{table}[t]
\centering
\caption{\textbf{Reference points of decouple and recouple modules.} 
}
\setlength{\tabcolsep}{5.0mm}
\begin{center}
    \begin{tabular}{c c c}
    \hline
     Points   & NDS/mAP  & mRR\\
    \hline
        2 & 70.9/68.3 & 77.6$\%$\\
        \rowcolor{gray!20}
        \textbf{4} & \textbf{71.5}/68.9 & \textbf{78.5}$\%$\\
        8 &  71.5/68.9  & 78.2$\%$\\
        12  & 71.4/\textbf{69.0} & 77.7$\%$ \\
    \hline
    \end{tabular}  
    \label{tab:ablations_points}
\end{center}
\end{table}

\begin{table}[t]
    \centering
    \caption{\textbf{Computational complexity comparison}. }
\setlength{\tabcolsep}{0.8mm}
\begin{tabular}{c|cc|cc}
\hline 
Model   & FLOPs↓ & FPS↑ & NDS/mAP & mRR \\
\hline

BEVFusion \cite{liu2023bevfusion}  & \textbf{115} GFLOPs & \textbf{4.0}& 71.0/68.5&  77.4$\%$\\
MetaBEV \cite{ge2023metabev} & 157 GFLOPs & 3.7& 71.3/68.6 & 77.9$\%$\\
Ours  & 140 GFLOPs & 3.9&  \textbf{72.0/69.5} &  \textbf{81.7$\%$}\\
\hline
\end{tabular}
    \label{table::model_size_time}
\end{table}

\begin{table}[t]
\centering
\caption{\textbf{Impact of pre-trained backbones.} 
}
\setlength{\tabcolsep}{2.5mm}
\begin{center}
    \begin{tabular}{c c c}
    \hline
     pre-trained backbones   & training epochs  & NDS/mAP\\
    \hline
        \checkmark & 6 & 72.0/69.5\\
        $\times$ & 6 & 67.8/62.1\\
        \rowcolor{gray!20}
        $\times$ & 24 & 71.7/69.2\\
        $\times$ & 48 & \textbf{72.5/69.8}\\
    \hline
    \end{tabular}  
    \label{tab:load_pre_train}
\end{center}
\end{table}


\subsection{Ablation studies of the network}

In this section, we conduct ablation studies to verify the effectiveness of the Multi-Modal Decouple and Recouple Encoder in our network. The metric mRR is computed on all types of data corruption. All experiments use BEVFusion as the base model.

\subsubsection{\textbf{Decouple and recouple modules}}

First, we verify the effectiveness of the two main modules in our network in Table \ref{tab:ablations_our_modules}. The results show that combining the two modules leads to better accuracy than using a single one. The modality decouple module extracts modality-invariant features and modality-specific features, in which modality-invariant features can be used for robust fusion under data corruption. And the modality recouple module further leverages useful information from corrupted modalities and invariant features and then adaptively fuses to create robust features.

\subsubsection{\textbf{Modality specific encoder in the decouple module}}

The comparison between (a) and (c) in Table \ref{tab:ablations_specific_encoder} shows that the performance gain also comes from our specific encoder. The comparison between (b) and (c) indicates that our encoder design is more robustness than the simple encoder design. The reason is that deformable attention can adaptively extract uncorrupted features, which allows the model to extract more useful information from corrupted modalities compared to simple encoder design.

\subsubsection{\textbf{Modality invariant encoder in the decouple module}}

Table \ref{tab:ablations_invariant} shows that combining all strategies of modality invariant encoder leads to better accuracy. Notably, row b shows that using only the $L_{Sim}$ and $L_{Diff}$ losses results in even worse performance than base model. This is because the encoder output may collapse to zero values, resulting in low loss values but no meaningful information. Row c demonstrates that introducing an auxiliary detection head ensures the output features are truly useful for 3D detection. With this head, $L_{Sim}$ forces the network to extract consistent features across modalities, while $L_{Diff}$ helps separate modality-invariant and modality-specific features.

\subsubsection{\textbf{Cross-modal recouple in the recouple module}}

Table \ref{tab:ablations_enhance} shows that using both the other modality and invariant features for enhancement achieves the best results. The other modality provides complementary information, while the invariant features offer robust information when any modality is corrupted.

\subsubsection{\textbf{Adaptive fusion in the recouple module}}

Table \ref{tab:ablations_moe_recouple}  shows that combining all strategies significantly improves robustness under corruptions. The entropy loss and different inputs for each expert make them more distinct. In contrast, giving all experts the same input reduces their distinct and affect the fusion results. Soft fusion strategy enables the model to adaptively recouple features. In contrast, using only the expert with the highest weight may also miss useful information from others.

\subsubsection{\textbf{Layers of our modules}}

We investigate the number of layers for our modality decouple module (MD) and modality recouple module (MR). Table \ref{tab:ablations_modules_layer} shows that MD and MR with a single layer perform more effectively than with multiple layers. It is because more parameters may lead to feature overfitting, which is not beneficial to the robustness of our model.

\subsubsection{\textbf{Reference points of our modules}}

We also investigate the number of reference key points selected in the deformable attention mechanism in our network. Table \ref{tab:ablations_points} verifies that using 4 key points achieves desired results and further increasing the number does not improve the accuracy. In addition, adding more key points may increase the computational cost and the chance of selecting corrupted features.

\subsubsection{\textbf{Impact of pre-trained backbones}}

Table \ref{tab:load_pre_train} shows that our model can get on-par performance when trained 24 epochs without loading pre-trained models, compared to training 6 epochs with pre-trained models. Training more epochs like 48 can further improve the performance.

\subsubsection{\textbf{Computational complexity}}

We test the computational complexity on an RTX 3090GPU with data loaded from a hard disk drive in Table \ref{table::model_size_time}. Compared to the base model BEVFusion, our method achieves significant improvements in accuracy and robustness without a noticeable increase in computational complexity. Compared with the recent robust model MetaBEV, our method achieves faster inference and higher accuracy. The reason is that each encoder in our model uses only one transformer layer, while MetaBEV uses six layers in total. In addition, our deformable attention is also more efficient than the standard Transformer in MetaBEV.

\begin{figure*}[t]
    \centering
    \includegraphics[width=18cm]{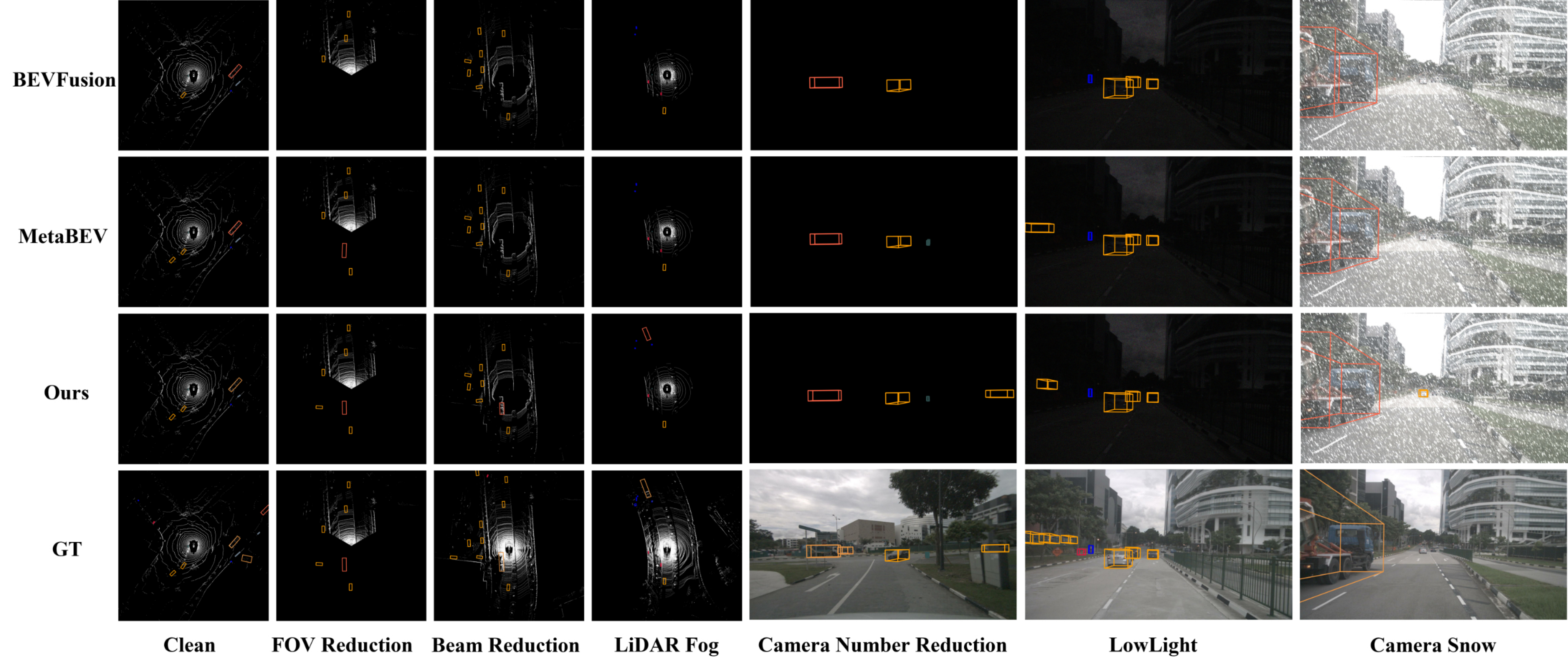}
    \caption{\textbf{Visual results on different types of corrupted data}. Our model detects more targets, while other baselines miss a few targets on corrupted data.
}
    \label{fig:detection_vis}
\end{figure*}

\begin{figure}[t]
    \centering
    \includegraphics[width=8cm]{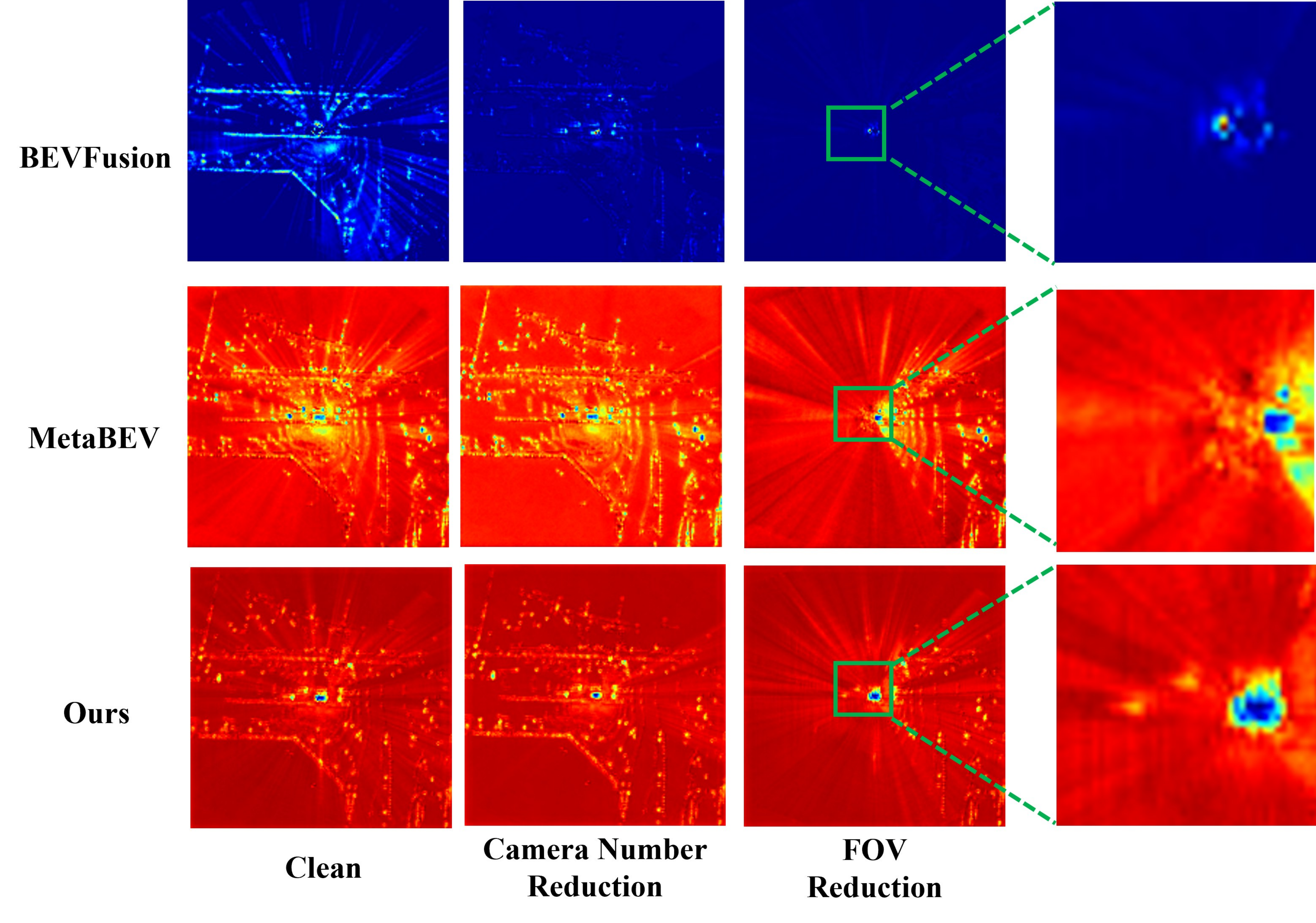}
    \caption{\textbf{Feature maps visualization}. The features of our model keep more consistent on clean data and corrupted data compared to the baselines.
}
    \label{fig:featmap_vis}
\end{figure}

\begin{figure}[t!]
    \centering
    \includegraphics[width=1.0\linewidth]{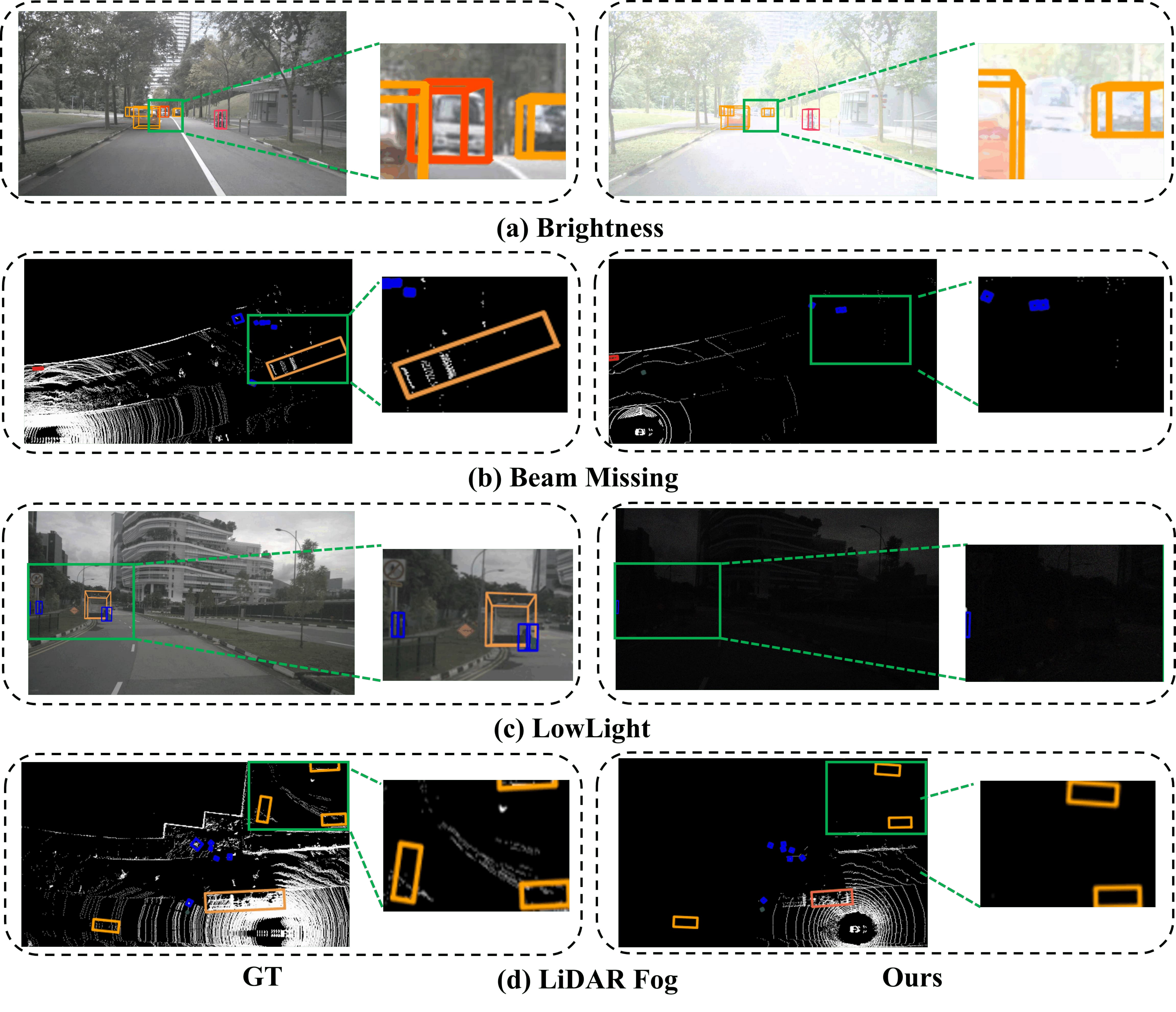}
    \caption{\textbf{Failure examples.}}
    \label{fig:bad_cases}
\end{figure}

\subsection{Visualization}

\subsubsection{\textbf{Feature maps visualization}}
\label{section_feat_map}

We analyze feature maps of our model and the two representative baselines MetaBEV and BEVFusion. 
Figure \ref{fig:featmap_vis} shows that BEVFusion well extracts the features on clean data, however, the overall system performance drops on corrupted data due to the limitation of its fusion strategy. MetaBEV well extracts the features on both clean data and corrupted data. However, it is activated in both foreground and background, which consequently leads to false positives. By comparison, our model focuses on foreground target areas and consistently extracts the features on all types of data corruption. Even LiDAR data is lost in the green box, our model is still activated on foreground target areas, while MetaBEV is completely inactivated. It further verifies the robustness of our model on data corruption in real-world applications.

\subsubsection{\textbf{Visual detection results}}
\label{section_vis_det}

Figure \ref{fig:detection_vis} visualizes the detection results of our model and the baselines on various types of data corruption. The results show that our model well detects more targets on all these types of data corruption, while the baselines often miss a few targets. The reason lies that our model effectively utilizes complementary information of two modalities when data corruption occur on LiDAR, camera, or both in real-world applications.

\subsubsection{\textbf{Failure examples}}

Though our model exhibits the best robustness in all types of data corruption, a few failure cases still occur in Figure \ref{fig:bad_cases} due to two reasons. On one hand, distant or small targets often have few useful information from both camera and LiDAR. Therefore, these targets are easily affected by data corruption (e.g., brightness or beam missing). On the other hand, some local regions from both camera and LiDAR are completely lost. Sever corruptions (e.g., fog) affect information from all modalities in some local regions and all targets in these regions cannot be detected.

\section{Conclusion}
\label{section_conclusion}

This paper introduces a Multi-Modal Decouple and Recouple network for robust multi-modal 3D object detection under data corruption of cameras, LiDAR, or both in real world. Extensive experiments verify that the model consistently achieves competitive results on both corrupted and clean data. The model does not require retraining or fine-tuning on specific types of data corruption, which is beneficial for applications in the real world. Future work will focus on more complex scenarios where LiDAR and camera may be corrupted by multiple types of data corruption simultaneously.


{
\small

\bibliographystyle{IEEEtran}


}


\end{document}